\documentclass{bvm} 
\begin{document}
\selectlanguage{english} 

\title{An Active Learning Pipeline for Biomedical Image Instance Segmentation with Minimal Human Intervention}

\titlerunning{An ActL Pipeline for Biomedical Image Segmentation}

\author{
	Shuo \lname{Zhao} \inst{1}, 
        Yu \lname{Zhou}  \inst{1,2},
	Jianxu \lname{Chen} \inst{1},
}

\authorrunning{Zhao, Zhou \& Chen}

\institute{
\inst{1} Leibniz-Institut für Analytische Wissenschaften - ISAS - e.V.\\
\inst{2} Faculty of Computer Science, Ruhr University Bochum
}

\email{jianxu.chen@isas.de}

\maketitle


\begin{abstract}
Biomedical image segmentation is critical for precise structure delineation and downstream analysis. Traditional methods often struggle with noisy data, while deep learning models such as U-Net have set new benchmarks in segmentation performance. nnU-Net further automates model configuration, making it adaptable across datasets without extensive tuning. However, it requires a substantial amount of annotated data for cross-validation, posing a challenge when only raw images but no labels are available. Large foundation models offer zero-shot generalizability, but may underperform on specific datasets with unique characteristics, limiting their direct use for analysis. This work addresses these bottlenecks by proposing a data-centric AI workflow that leverages active learning and pseudo-labeling to combine the strengths of traditional neural networks and large foundation models while minimizing human intervention. The pipeline starts by generating pseudo-labels from a foundation model, which are then used for nnU-Net's self-configuration. Subsequently, a representative core-set is selected for minimal manual annotation, enabling effective fine-tuning of the nnU-Net model. This approach significantly reduces the need for manual annotations while maintaining competitive performance, providing an accessible solution for biomedical researchers to apply state-of-the-art AI techniques in their segmentation tasks. The code is available at \url{https://github.com/MMV-Lab/AL_BioMed_img_seg}.
\end{abstract}

\section{Introduction}

Image segmentation is crucial in biomedical image analysis, as it allows precise delineation of structures, which could be used for downstream quantification and modeling. The field of biomedical image segmentation has evolved significantly with deep learning techniques. Although traditional methods such as thresholding or graph-based methods were foundational, they often yield much worse performance on challenging tasks (e.g., noisy data) compared to deep learning-based approaches such as U-Net \cite{3627-01}.

Traditional neural networks such as nnU-Net further improved this by automating model configuration for different segmentation tasks \cite{3627-02}, which has become a strong baseline method in segmentation benchmarks.  nnU-Net can adapt to various biomedical imaging datasets without extensive manual tuning, effectively overcoming the challenges associated with manual network design and configuration. Empirical selection requires a complete training set or sufficient data for five-fold cross-validation. However, obtaining labeled data for nnU-Net, especially in 3D image analysis, is time-consuming and requires knowledge and expertise. In practice, data often has no labels.

Unsupervised active learning advances core-set selection for efficient training and lower annotation costs but faces challenges in feature selection and initialization. \cite{3627-15}.

\begin{figure}[b] 
    \centering 
    \includegraphics[width=\textwidth]{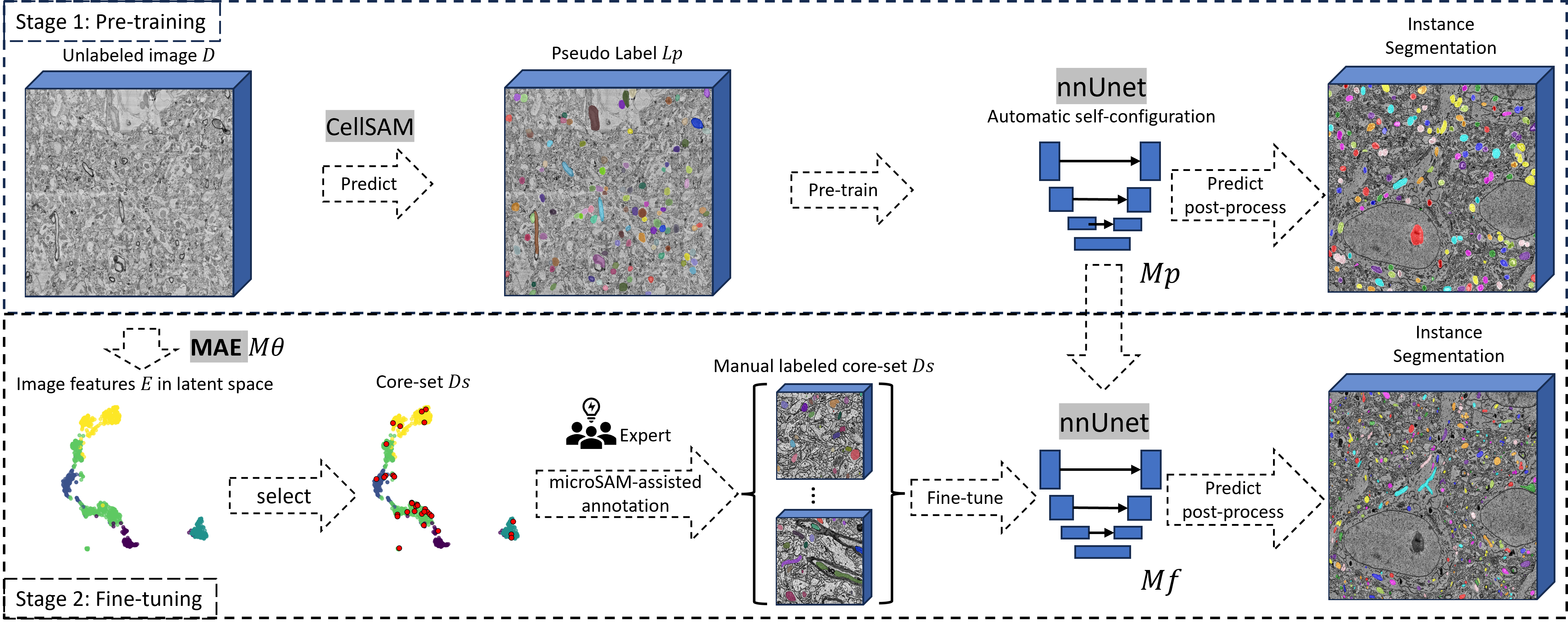} 
    \caption{An active learning pipeline for 3D unlabeled biomedical image instance segmentation.} This pipeline consists of two main stages: Pre-training and Fine-tuning. Stage 1: Pre-training. Pseudo-labels were generated from CellSAM \cite{3627-06} for unlabeled raw images, forming a training set for nnU-Net to run self-configuration and pre-training. Stage 2: Fine-tuning. Unlabeled images were mapped into a latent space to select a representative core-set \cite{3627-11}, using features from a pre-trained foundation model MAE \cite{3627-12} via self-supervised learning. The pre-trained nnUnet model was fine-tuned by using minimal microSAM-assisted annotations. 
    \label{3627-fig-01}
\end{figure}

Meanwhile, different network architectures, such as transformer-based models and hybrid models combining convolutional neural networks (CNNs) with transformers, have also been proposed to improve segmentation by capturing long-range dependencies in images \cite{3627-03}. Recent developments in large foundation models, such as SAM \cite{3627-04}, SAM2 \cite{3627-05}, cellSAM \cite{3627-06}, microSAM \cite{3627-07}, medSAM \cite{3627-08}, have further advanced the field toward generalist models for biomedical image segmentation. However, large foundation models often struggle with specific datasets, such as unique tissue environments or drug-treated cells, limiting their effectiveness in direct downstream analysis.

We address the two bottlenecks with an active learning pipeline based on the idea of data-centric AI to solve these two bottlenecks, which focuses on data quality and diversity rather than model complexity \cite{3627-09}. This pipeline combines traditional neural networks and large foundation models with minimal human input. A 3D mitochondrial segmentation \cite{3627-10} was used as an example. Pseudo-labels generated from a foundation model for raw images can be used to create a training set for self-configuration and pre-training. This approach enables nnUNet to work with unlabeled images. A core-set is a small, representative subset that approximates the properties of the full dataset. Then a core-set was selected for manual annotation and fine-tuning the pre-rained nnUnet, leading to a robust model with minimal manual effort. It is worth mentioning that this paper focuses on an active learning pipeline to address key bottlenecks in traditional neural networks and large foundation models. So we focus on demonstrating and validating the effectiveness of this pipeline rather than competing on leaderboards.

\section{Materials and methods}

\subsection{Methods} 

The proposed pipeline (Fig.~\ref{3627-fig-01}) consists of two main stages: pre-training and fine-tuning.

\emph{Stage 1: Pre-training}. To enable nnU-Net $M_{p}$
 to run self-configuration and pre-training, avoiding manual design, pseudo-labels $L_{p}$ were generated from CellSAM \cite{3627-06} for unlabeled raw images $D$. Then $D$ and $L_{p}$ formed a train set for nnU-Net $M_{p}$. 
 
 \emph{Stage 2: Fine-tuning}. The selected core-set $D_{s}$ and its manual label were used to fine-tune the pre-trained nnU-Net $M_{p}$. We selected raw images based on their latent space features. An MAE model $M_{\theta}$ using U-net architecture was trained with unlabeled images $D$ via self-supervised learning. According to $E = M_{\theta}(D)$, $D$ were mapped into a latent space as the features $E$. Pre-trained MAE models excel in scalability, speed, and feature extraction, outperforming supervised models \cite{3627-14}. According to $d = 1 - E \cdot E^T$, a cosine distance matrix $d$ was computed between features $E$. 
 
 Unlabeled images $D$ consists of a selected core-set $D_{s}$ \cite{3627-11} and unselected set $D_{u}$. The core-set selection algorithm, initialization, $k$ ($k = 3$) samples were selected randomly from $D_{u}$. Then, each sample $S$ in $D_{u}$ was traversed, find the minimum distance $d_{\text{min}}(S, D_s)$ for each $S$ to $D_{s}$. Then the sample $S$ in $D_{u}$ with the maximum $d_{\text{min}}$ was selected to $D_{s}$. We repeated this process, and finally get a core-set $D_{s}$ of size $N$. 
 
 The unlable selected core-set $D_{s}$ was checked and annotated by experts with minimal manual effort, assisted by microSAM \cite{3627-07}. Then $D_{s}$ and manual labels were used to fine-tune the nnU-Net $M_{p}$, resulting in $M_{f}$. Connected components were applied to make instance segmentation as in MitoEM Challenge \cite{3627-10}.


\subsection{Validation on the 3D unlabeled dataset MitoEM} 
\label{3627-Materials and Methods}
The MitoEM Grand Challenge \cite{3627-10} is an international competition focused on segmenting mitochondria in electron microscopy images. The 3D MitoEM dataset \cite{3627-13} contains two high-resolution volumes, each measuring $30 \times 30 \times 30$ $\mu$m with voxel dimensions of $1000 \times 4096 \times 4096$ at $30 \times 8 \times 8$ nm resolution. Each volume is divided into training (400 slices), validation (100 slices), and test (500 slices) sets. \emph{Although it has ground truth (GT), we assume no GT is available for training, as in real-world tasks.}

\section{Experiment} 

All experiments were run on 8 NVIDIA A100 GPUs (40GB). For fair experiments, we simply adopt the hyperparameters in the original nnUnet codebase for training. 

For dataset preprocessing, train and test sets were padded and cropped into 1024 patches ($32 \times 512 \times 512$). 2D pseudo-labels were generated by CellSAM and combined into 3D patch labels. nnU-Net was pre-trained and fine-tuned automatically. MAE models were trained on 1000 raw 2D images with a 0.75 mask ratio for 400 epochs. Then the pre-trained encoders were used to extract image features. Patches were selected by the core-set selection algorithm ($k = 3$). Core-sets were annotated by microSAM with the \texttt{vit\_b} model. Fine-tuned models segmented the test set, with post-processing and evaluation following the MitoEM Challenge \cite{3627-10}. For ablation experiments, we compared nnU-Net models trained on the core-set with those pre-trained on pseudo-labels. We also assessed the impact of various core-set sizes, from 8 to 1024 samples.

\begin{table}[t]
    \caption{Evaluation metrics for fine-tuning with different core-set sizes. Where the 1024 core-set represents the full manually labeled set. Each double row displays the score and percentage of the full set score. \emph{Bold} text highlights when the score first surpasses 90\% of the full set performance.}
    \label{3627-tab-active_learning_metrics} 
    \centering
    \begin{tabular*}{\textwidth}{l@{\extracolsep{\fill}}ccccccccc}
        \hline
        \emph{Core-set Size} & \emph{0} & \emph{8} & \emph{16} & \emph{32} & \emph{64} & \emph{128} & \emph{256} & \emph{512} & \emph{1024} \\
        \emph{Percentage (\%)} & \emph{(0.00)} & \emph{(0.78)} & \emph{(1.56)} & \emph{(3.13)} & \emph{(6.25)} & \emph{(12.50)} & \emph{(25.00)} & \emph{(50.00)} & \emph{(100.00)} \\
        \hline

        \emph{F1 Score} & 0.4025 & 0.4462 & 0.5297 & 0.5630 & \emph{0.5884} & 0.6003 & 0.6011 & 0.6294 & 0.6350 \\
        \emph{Percentage (\%)} & 63.38 & 70.26 & 83.41 & 88.66 & \emph{92.65} & 94.53 & 94.65 & 99.12 & 100.00 \\
        \emph{Accuracy} & 0.2521 & 0.2882 & 0.3603 & 0.3918 & 0.4170 & \emph{0.4291} & 0.4298 & 0.4593 & 0.4652 \\
        \emph{Percentage (\%)} & 54.19 & 61.95 & 77.45 & 84.22 & 89.63 & \emph{92.23} & 92.39 & 98.72 & 100.00 \\
        \emph{Panoptic} & 0.3628 & 0.3962 & 0.4723 & 0.5038 & \emph{0.5284} & 0.5405 & 0.5421 & 0.5689 & 0.5750 \\
        \emph{Percentage (\%)} & 63.09 & 68.90 & 82.13 & 87.61 & \emph{91.89} & 93.99 & 94.26 & 98.93 & 100.00 \\
        \emph{Precision} & 0.2824 & 0.3300 & 0.4181 & 0.4599 & 0.4926 & \emph{0.5045} & 0.5058 & 0.5411 & 0.5554 \\
        \emph{Percentage (\%)} & 50.85 & 59.42 & 75.29 & 82.81 & 88.70 & \emph{90.85} & 91.08 & 97.43 & 100.00 \\
        \hline
    \end{tabular*}
\end{table}

\begin{figure}[b]
	\centering 
	\setlength{\figwidth}{0.14\textwidth}
	\begin{subfigure}{\figwidth}
		\includegraphics[width=\textwidth]{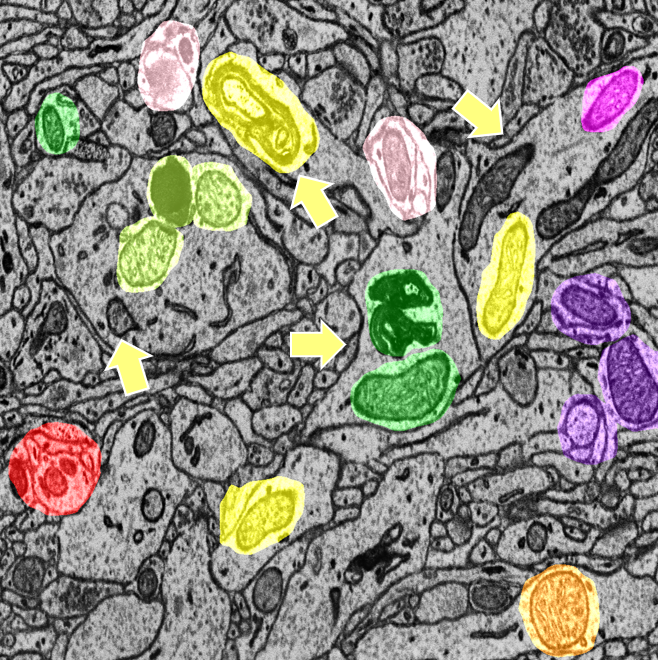}
		\caption{}
		\label{3627-fig-02-a}
	\end{subfigure}
	\hfill	
        \begin{subfigure}{\figwidth}
		\includegraphics[width=\textwidth]{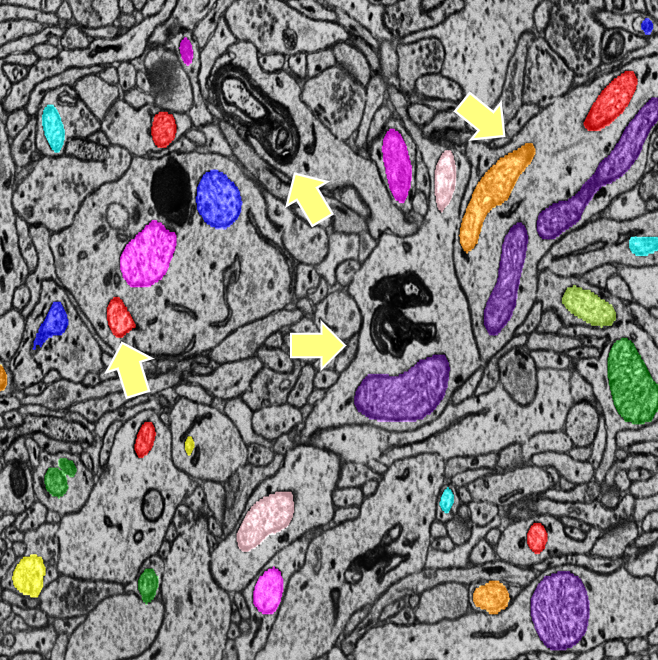}
		\caption{}
		\label{3627-fig-02-b}
	\end{subfigure}
	\hfill	
	\begin{subfigure}{\figwidth}
		\includegraphics[width=\textwidth]{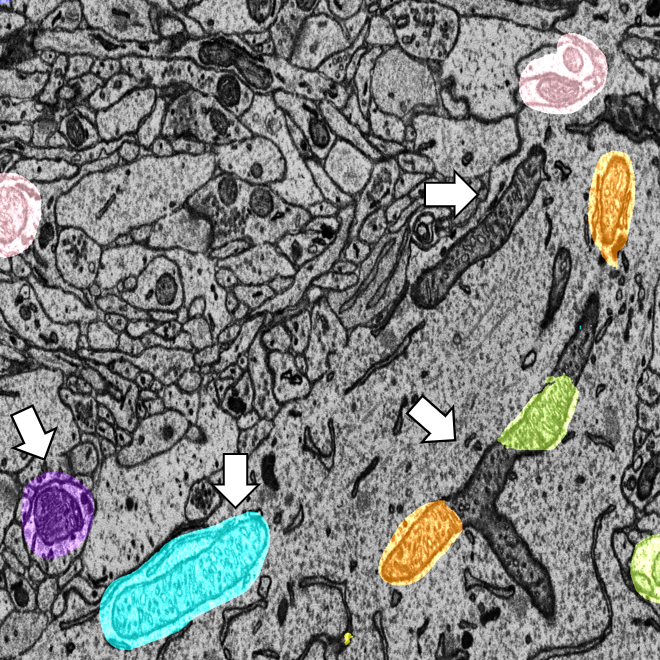}
		\caption{}
		\label{3627-fig-02-c}
	\end{subfigure}
	\hfill	
	\begin{subfigure}{\figwidth}
		\includegraphics[width=\textwidth]{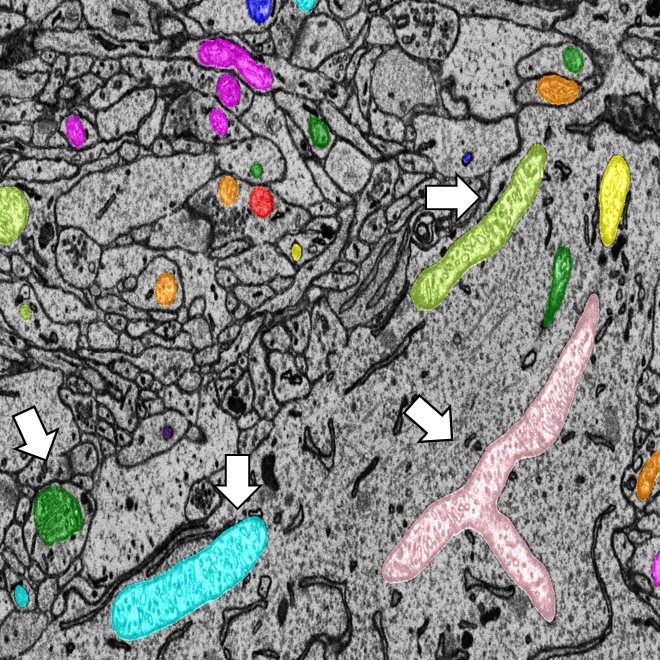}
		\caption{}
		\label{3627-fig-02-d}
	\end{subfigure}
	\hfill	
	\begin{subfigure}{\figwidth}
		\includegraphics[width=\textwidth]{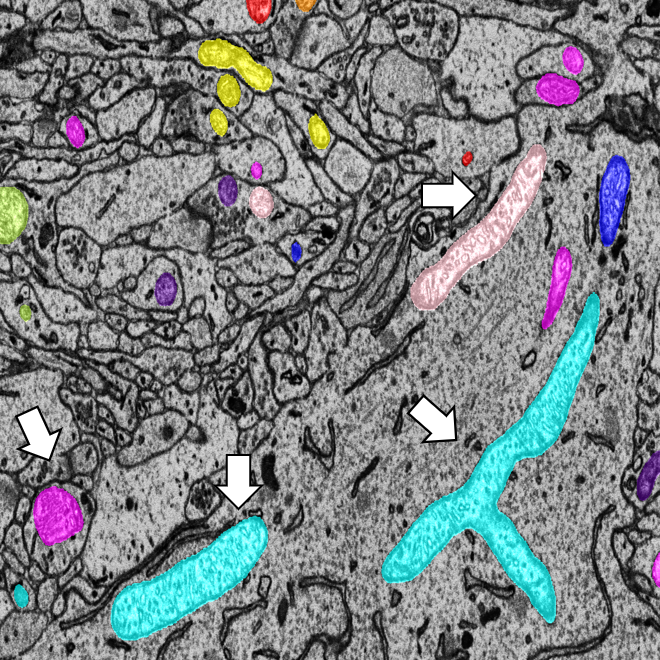}
		\caption{}
		\label{3627-fig-02-e}
	\end{subfigure}
	\hfill
	\begin{subfigure}{\figwidth}
		\includegraphics[width=\textwidth]{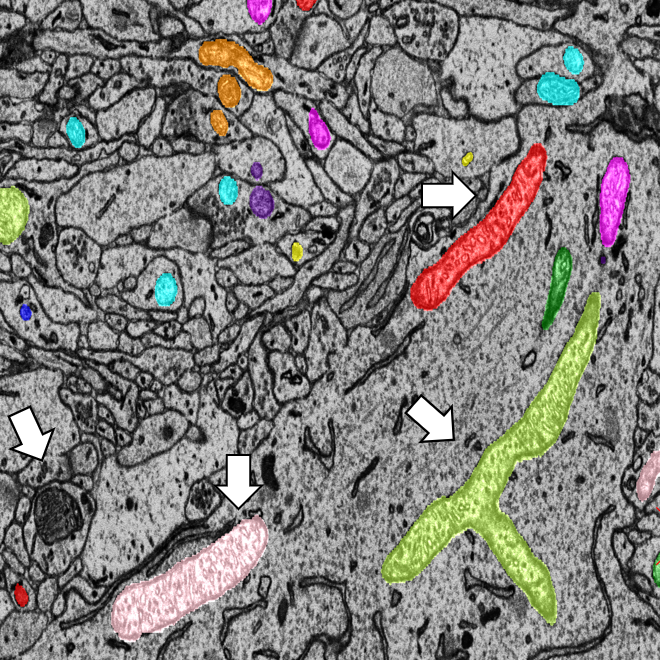}
		\caption{}
		\label{3627-fig-02-f}
	\end{subfigure}
	\caption{Labels for the train set: (a) pseudo-labels from cellSAM and (b) manual labels. Segmentations for the test set: (c) pre-trained with pseudo-labels, (d) fine-tuned with 12.5\% manual labels, (e) fine-tuned with 100\% manual labels, and (f) GT. Yellow arrows in (a) and (b) show differences between pseudo and manual labels, while white arrows in (c)–(f) highlight discrepancies between segmentations and the GT.}
	\label{3627-fig-02}
\end{figure}

\section{Results} 

\begin{SCfigure}[10][h]
    \centering 
    \includegraphics[width= 0.45 \textwidth]{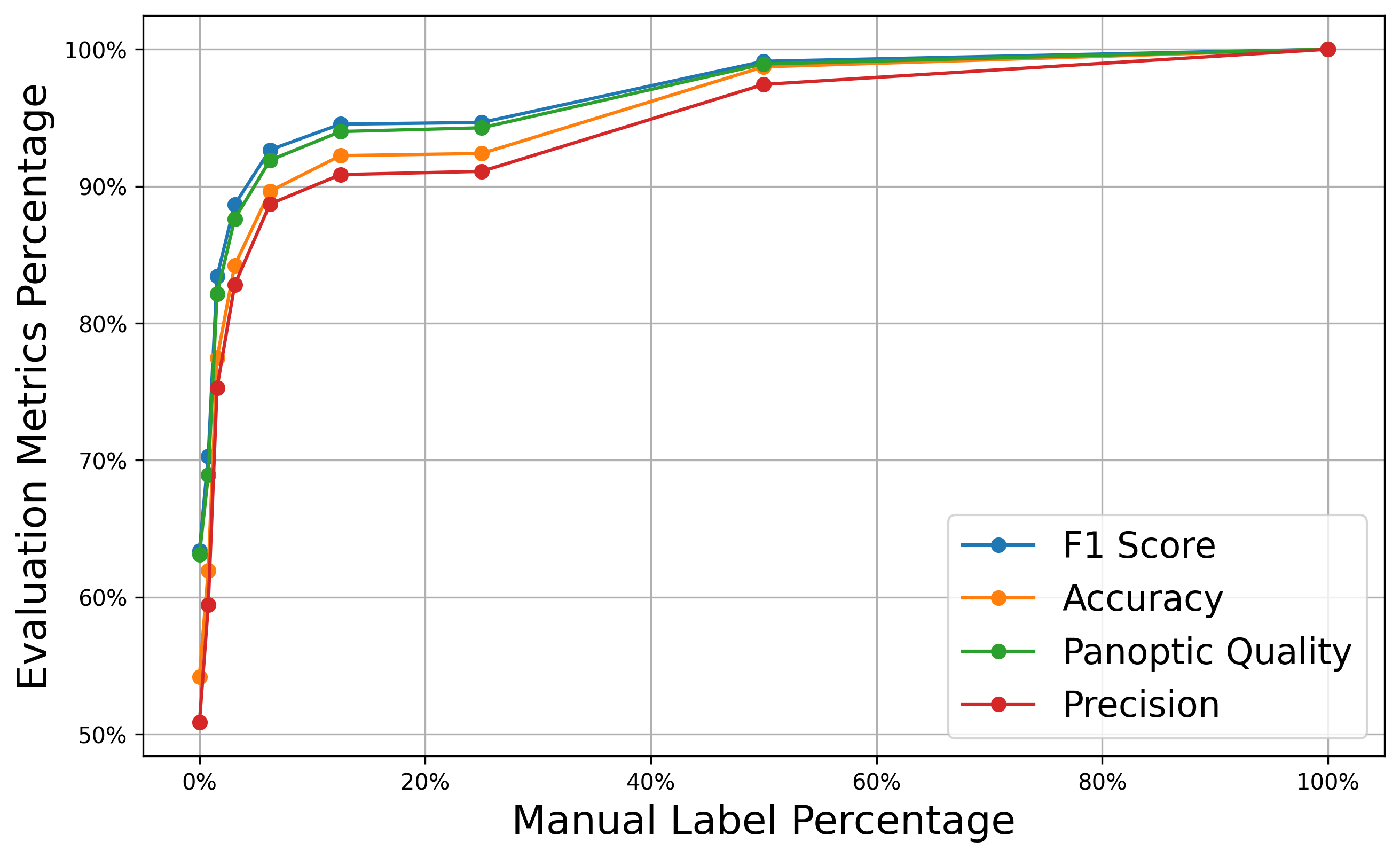} 
    \caption{Instance segmentation metrics: The X-axis shows manual annotation percentage, and the Y-axis shows performance relative to full labels. As manual annotations increase, the model's performance improves. At about 12.5\% annotations, the model achieves 90\% of the full annotation performance.} 
    \label{3627-fig-04} 
\end{SCfigure}

\begin{SCfigure}[10][h]
    \centering 
    \includegraphics[width= 0.45 \textwidth]{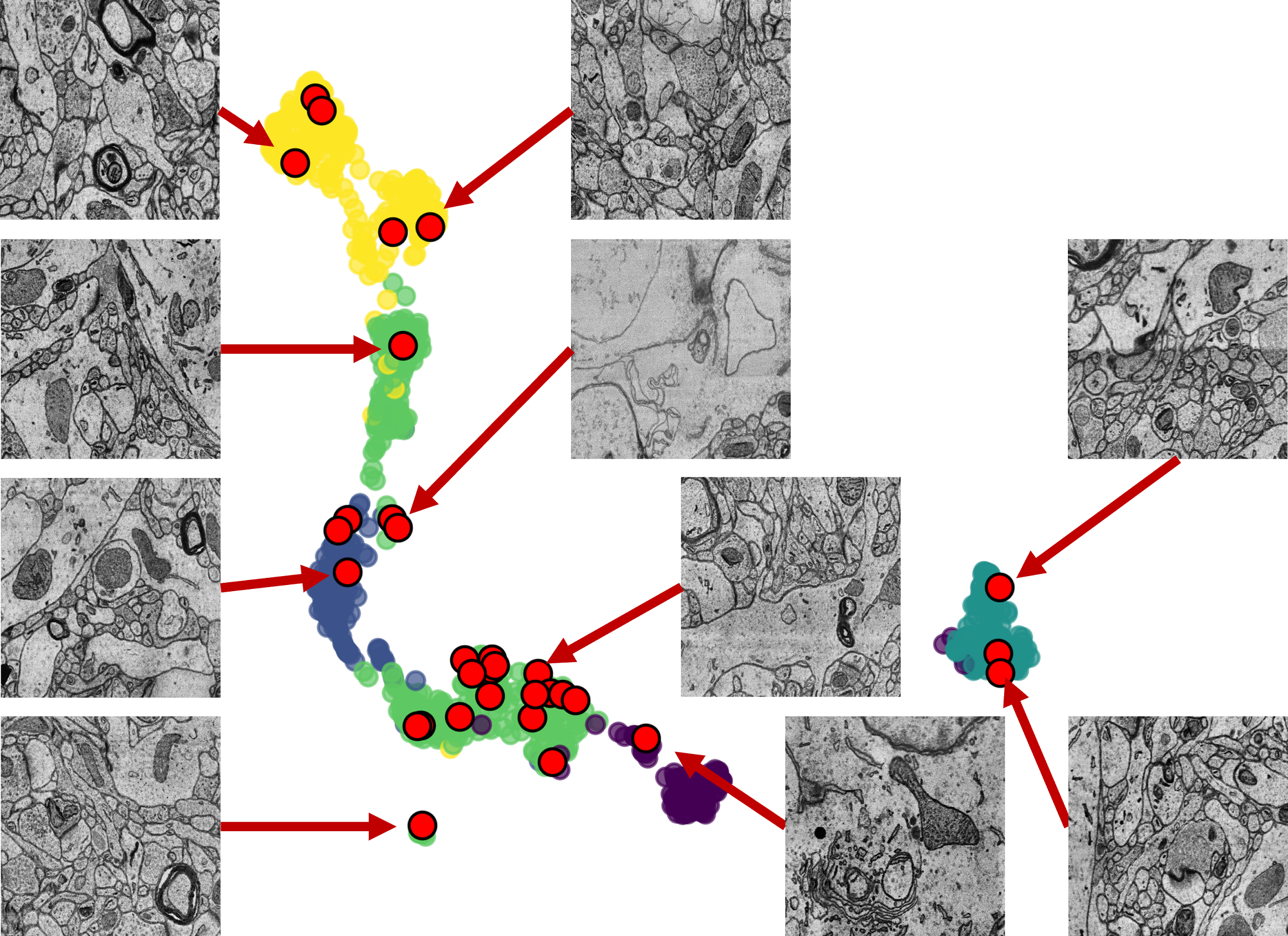} 
    \caption{Examples of patches in the selected core-set. The plot visualizes the latent features of all patches in latent space after UMAP dimensionality reduction. Red dots represent the selected image patches. The selected patches cover a wide range but have different features. Compared with random selection, the selected core-set is both representative and diverse, with high information density.} 
    \label{3627-fig-03} 
\end{SCfigure}

We observed that \emph{simply relying on pseudo-labels from large foundation models may be insufficient}. As shown in Fig. ~\ref{3627-fig-02}, pseudo-labels (Fig. ~\ref{3627-fig-02-a}) closely resemble the manual labels in (Fig. ~\ref{3627-fig-02-b}). However, yellow arrows highlight labeling errors, e.g., imprecise boundaries and false negatives. The white arrows indicate that, compared to GT (Fig. ~\ref{3627-fig-02-f}), the segmentation after pre-training (Fig. ~\ref{3627-fig-02-c}) detects mitochondria but has inaccuracies. \emph{The selected core-sets are reasonable and effective}. Fig. ~\ref{3627-fig-03} shows the latent space distribution of image features, with red points marking diverse, representative patches that enhance information density. \emph{Fine-tuning with the selected core-set is effective}. As shown in Fig. ~\ref{3627-fig-02}, fine-tuning with 12.5\% and 100\% manual labels (Fig. ~\ref{3627-fig-02-d}, Fig. ~\ref{3627-fig-02-e}) improves segmentations closer to the GT (Fig. ~\ref{3627-fig-02-f}). Fig. ~\ref{3627-fig-04} and Tab. ~\ref{3627-tab-active_learning_metrics} show fine-tuning results with different core-set sizes. Tab.~\ref{3627-tab-active_learning_metrics} highlights that F1 score and Panoptic metrics surpassed 90\% at 6.25\% (64 samples), and Accuracy and Precision at 12.5\% (128 samples). Fine-tuning surpasses pseudo-label-only pre-training (core-set = 0). \emph{nnU-Net leverages pseudo-label pre-training for auto-configuration.}. As shown in Tab.\ref{3627-tab-active_learning_metrics},
core-set fine-tuned models consistently outperformed pseudo-label-only pre-trained models. \emph{Core-set selection outperforms random selection}. Tab.\ref{3627-tab-core-set-vs-random} shows that with 128 (12.5\%) samples, the core-set improved F1 score, accuracy, panoptic, and precision, making core-set selection with pre-training the most effective for fine-tuning nnU-Net.

\begin{table}[t]
    \caption{Evaluation metrics for Core-set and Random selection. A train set of 128 (12.5\%) samples was selected to compare the performance of the core-set selection and randomly selection. W/ indicates nnU-Net was pre-trained on pseudo-labels before fine-tuning, while w/o means it was trained directly on the selected set without pre-training.}
    \label{3627-tab-core-set-vs-random}
    \centering
    \begin{tabular*}{\textwidth}{l@{\extracolsep{\fill}}ccccc}
        \hline
        \emph{} & \emph{Pre-trained} & \emph{F1 score} & \emph{Accuracy} & \emph{Panoptic} & \emph{Precision} \\
        \hline
        \emph{Core-set} & \emph{w/}  & 0.6003 & 0.4291 & 0.5405 & 0.5045 \\
        \emph{Core-set} & \emph{w/o} & 0.5939 & 0.4225 & 0.5342 & 0.4953 \\
        \emph{Random}   & \emph{w/}  & 0.5773 & 0.4058 & 0.5173 & 0.4697 \\
        \emph{Random}   & \emph{w/o} & 0.5793 & 0.4076 & 0.5183 & 0.4732 \\
        \hline
    \end{tabular*}
\end{table}

\section{Discussion and conclusion} 

Pseudo-labels from CellSAM enable nnU-Net training without manual annotations. MicroSAM refines these labels semi-automatically, reducing effort. This pipeline integrates foundation models with traditional networks for efficient raw data use. Core-set fine-tuning surpasses pseudo-label pre-training and random selection.

In summary, this work presents an active learning workflow \emph{addressing two major pain-points in existing methods:} (1) nnUnet requires considerable ground truth, but usually not available in real applications, for auto-configuration, (2) foundation models cannot always provide completely satisfactory results off-the-shelf. The proposed framework seamlessly streamlines two powerful methods with minimal human intervention. Future work will enable continuous learning for dynamic datasets.

\begin{disclaimer}
\texttt{The authors declare that there is no conflict of interest regarding the publication of this paper.} 
\end{disclaimer}

\begin{acknowledgement}
\texttt{This work is supported by the by the Federal Ministry of Education and Research (BMBF) in Germany under the funding reference 161L0272, and also supported by the Ministry of Culture and Science (MKW) of the State of North Rhine-Westphalia}.	
\end{acknowledgement}

\printbibliography

\end{document}